\documentclass{article}

\usepackage[final]{neurips_2024}

\usepackage[utf8]{inputenc}
\usepackage[T1]{fontenc}
\usepackage{hyperref}
\usepackage{url}
\usepackage{booktabs}
\usepackage{amsfonts}
\usepackage{nicefrac}
\usepackage{microtype}
\usepackage{xcolor}
\usepackage{graphicx}
\usepackage{amsmath}
\usepackage{amssymb}
\usepackage{multirow}
\usepackage{subcaption}
\usepackage{algorithm}
\usepackage{algorithmic}

\title{Do Large Language Models Know What They Don't Know? \\ Evaluating Epistemic Calibration via Prediction Markets}

\author{
  Lukas Nel\\
  Lotus AI\\
  \texttt{lukas@lotus.ai}
}

\begin{document}

\maketitle

\begin{abstract}
A well-calibrated model should express confidence that matches its actual accuracy---when it claims 80\% confidence, it should be correct 80\% of the time. While large language models (LLMs) have achieved remarkable performance across diverse tasks, their epistemic calibration remains poorly understood. We introduce \textbf{KalshiBench}, a benchmark of 300 prediction market questions from Kalshi, a CFTC-regulated exchange, with verifiable real-world outcomes occurring after model training cutoffs. Unlike traditional benchmarks measuring accuracy on static knowledge, KalshiBench evaluates whether models can appropriately quantify uncertainty about genuinely unknown future events. We evaluate five frontier models---Claude Opus 4.5, GPT-5.2, DeepSeek-V3.2, Qwen3-235B, and Kimi-K2---and find \textbf{systematic overconfidence across all models}. Even the best-calibrated model (Claude Opus 4.5, ECE=0.120) shows substantial calibration errors, while reasoning-enhanced models like GPT-5.2-XHigh exhibit \emph{worse} calibration (ECE=0.395) despite comparable accuracy. Critically, only one model achieves a positive Brier Skill Score, indicating most models perform worse than simply predicting base rates. Our findings suggest that scaling and enhanced reasoning do not automatically confer calibration benefits, highlighting epistemic calibration as a distinct capability requiring targeted development.
\end{abstract}

\section{Introduction}

The deployment of large language models in high-stakes domains---medical diagnosis, legal reasoning, financial forecasting---demands not only accuracy but also \emph{calibrated uncertainty}. A model claiming ``90\% confidence'' in a diagnosis should be correct approximately 90\% of the time on similar cases. Poor calibration manifests as dangerous overconfidence (trusting wrong answers) or unnecessary underconfidence (ignoring correct ones), fundamentally limiting the utility of model predictions for decision-making under uncertainty \citep{guo2017calibration}.

Despite extensive work on LLM capabilities, epistemic calibration---the alignment between expressed confidence and actual accuracy---remains understudied. Existing evaluations face two critical limitations:

\textbf{(1) Static knowledge contamination.} Traditional benchmarks assess models on questions whose answers existed during training. A model may appear ``calibrated'' simply by having memorized facts with appropriate confidence, rather than genuinely reasoning about uncertainty.

\textbf{(2) Lack of verifiable ground truth.} Many calibration studies rely on human judgments or synthetic datasets, introducing noise and potential biases in ground truth labels.

We address both limitations through \textbf{KalshiBench}, a benchmark leveraging prediction markets---specifically Kalshi, a CFTC-regulated exchange where contracts resolve to verifiable real-world outcomes. By temporally filtering questions to those resolving \emph{after} model training cutoffs, we ensure models cannot have memorized outcomes, providing a clean signal for epistemic calibration.

\begin{figure}[t]
    \centering
    \begin{minipage}{0.48\textwidth}
        \centering
        \small
        \begin{tabular}{lcc}
        \toprule
        \textbf{Model} & \textbf{Accuracy} & \textbf{ECE} $\downarrow$ \\
        \midrule
        Claude Opus 4.5 & 69.3\% & \textbf{0.120} \\
        Kimi-K2 & 67.1\% & 0.298 \\
        Qwen3-235B & 65.7\% & 0.297 \\
        GPT-5.2-XHigh & 65.3\% & 0.395 \\
        DeepSeek-V3.2 & 64.3\% & 0.284 \\
        \bottomrule
        \end{tabular}
    \end{minipage}
    \hfill
    \begin{minipage}{0.48\textwidth}
        \centering
        \small
        \textbf{Key Finding:} All models exhibit systematic overconfidence. The gap between confidence and accuracy widens dramatically at high confidence levels, with models averaging 27\% error rate even when expressing $>$90\% confidence.
    \end{minipage}
    \caption{Summary of main results. While accuracy varies modestly (64-69\%), calibration error varies dramatically (3$\times$ range). Reasoning enhancements (GPT-5.2-XHigh) worsen rather than improve calibration.}
    \label{fig:summary}
\end{figure}

Our contributions are:

\begin{enumerate}
    \item \textbf{KalshiBench}: A temporally-filtered benchmark of 300 prediction market questions spanning 13 categories with verified ground truth outcomes, designed for rigorous calibration evaluation.
    
    \item \textbf{Comprehensive evaluation}: We assess five frontier models across classification (accuracy, F1) and calibration (Brier score, ECE, reliability diagrams) metrics, revealing systematic patterns.
    
    \item \textbf{Novel findings}: We demonstrate that (a) all current frontier models are overconfident, (b) reasoning enhancements degrade calibration, (c) only one model beats the base-rate baseline, and (d) calibration and accuracy are largely decoupled.
\end{enumerate}

\section{Related Work}

\paragraph{Calibration in Neural Networks.}
Calibration has been extensively studied in classification settings \citep{guo2017calibration, minderer2021revisiting}. Modern deep networks are known to be overconfident \citep{guo2017calibration}, with various post-hoc calibration methods proposed including temperature scaling \citep{guo2017calibration}, Platt scaling \citep{platt1999probabilistic}, and isotonic regression \citep{zadrozny2002transforming}. However, these methods assume access to held-out calibration data and primarily address discriminative rather than generative models.

\paragraph{LLM Uncertainty Quantification.}
Prior work on LLM calibration has examined confidence elicitation through verbalized probabilities \citep{tian2023just, xiong2024can}, multiple sampling \citep{wang2023selfconsistency}, and logit-based approaches \citep{kadavath2022language}. \citet{kadavath2022language} found that larger models show improved calibration on factual questions, while \citet{tian2023just} demonstrated that verbalized confidence often diverges from token probabilities. Recent work has explored calibration in specific domains including medical question-answering \citep{singhal2023large} and mathematical reasoning \citep{lightman2023lets}.

\paragraph{Forecasting and Prediction Markets.}
Prediction markets aggregate collective intelligence to forecast uncertain events \citep{arrow2008promise, wolfers2004prediction}. Superforecasters demonstrate that calibration is a learnable skill \citep{tetlock2015superforecasting}. Recent work has begun exploring LLMs as forecasters \citep{zou2022forecasting, halawi2024approaching}. Most relevant to our work, ForecastBench \citep{karger2024forecastbench} introduced a dynamic benchmark evaluating ML forecasting on 1,000 automatically-updated questions, finding that expert human forecasters significantly outperform the best LLMs. However, ForecastBench focuses primarily on accuracy rather than calibration metrics.

\paragraph{Distinction from Prior Work.}
Unlike existing benchmarks that assess calibration on static knowledge questions, KalshiBench uses temporally-filtered prediction market questions with verified post-training outcomes, eliminating knowledge contamination and providing clean calibration signals. Compared to ForecastBench, we focus specifically on \emph{calibration} rather than raw forecasting accuracy, providing detailed analysis of reliability diagrams, overconfidence rates, and the relationship between confidence and correctness.

\section{KalshiBench Dataset}

\subsection{Data Source and Collection}

KalshiBench sources questions from Kalshi\footnote{\url{https://kalshi.com}}, a CFTC-regulated prediction market exchange operating in the United States. Unlike informal forecasting platforms, Kalshi contracts have legally-binding resolution criteria, ensuring unambiguous ground truth. The full KalshiBench dataset contains \textbf{1,531 cleaned, deduplicated prediction market questions} spanning from September 2021 to November 2025 across 16 categories, with a 42\%/58\% yes/no class split.

For our evaluation, we apply temporal filtering based on model knowledge cutoffs and randomly sample \textbf{300 questions} (random seed 42) from the filtered set. This sample size balances computational cost against statistical power, and exceeds the 200-question evaluation used in ForecastBench \citep{karger2024forecastbench}.

\subsection{Temporal Filtering}

To ensure models cannot have memorized outcomes, we apply strict temporal filtering based on model knowledge cutoffs:

\begin{equation}
    \mathcal{D}_{\text{filtered}} = \{(q, y) \in \mathcal{D} : t_{\text{close}}(q) > \max_{m \in \mathcal{M}} t_{\text{cutoff}}(m)\}
\end{equation}

where $t_{\text{close}}(q)$ is the resolution time of question $q$, $t_{\text{cutoff}}(m)$ is the knowledge cutoff of model $m$, and $\mathcal{M}$ is the set of evaluated models. For our evaluation, the effective cutoff is October 1, 2025 (the latest among all models).

\subsection{Dataset Statistics}

\begin{table}[h]
\centering
\caption{KalshiBench dataset statistics. The full dataset contains 1,531 questions; we evaluate on a temporally-filtered sample of 300 questions (seed=42) resolving after October 1, 2025.}
\label{tab:dataset_stats}
\begin{tabular}{ll|ll}
\toprule
\textbf{Full Dataset} & \textbf{Value} & \textbf{Evaluation Sample} & \textbf{Value} \\
\midrule
Total Questions & 1,531 & Sampled Questions & 300 \\
Categories & 16 & Categories (in sample) & 13 \\
Date Range & 2021-09 to 2025-11 & Date Range & 2025-10 to 2025-11 \\
Yes Rate & 42.3\% & Yes Rate & 40.0\% \\
Temporal Span & 1,537 days & Temporal Span & 46 days \\
\bottomrule
\end{tabular}
\end{table}

\begin{table}[h]
\centering
\caption{Category distribution in KalshiBench. Sports and Politics dominate, but all major forecasting domains are represented.}
\label{tab:categories}
\small
\begin{tabular}{lrrr|lrrr}
\toprule
\textbf{Category} & \textbf{N} & \textbf{\%} & \textbf{Yes\%} & \textbf{Category} & \textbf{N} & \textbf{\%} & \textbf{Yes\%} \\
\midrule
Sports & 83 & 27.7 & 34.9 & Crypto & 11 & 3.7 & 27.3 \\
Politics & 55 & 18.3 & 52.7 & Climate/Weather & 9 & 3.0 & 33.3 \\
Entertainment & 47 & 15.7 & 36.2 & Financials & 8 & 2.7 & 12.5 \\
Companies & 30 & 10.0 & 60.0 & World & 6 & 2.0 & 50.0 \\
Elections & 24 & 8.0 & 20.8 & Economics & 4 & 1.3 & 50.0 \\
Mentions & 19 & 6.3 & 36.8 & Social & 3 & 1.0 & 66.7 \\
\bottomrule
\end{tabular}
\end{table}

\subsection{Deduplication and Quality Control}

Raw prediction market data contains redundant questions (e.g., daily instances of recurring markets). We limit to 2 questions per series ticker to preserve diversity while reducing redundancy. All questions include detailed resolution criteria in the description field, ensuring unambiguous ground truth.

\section{Methodology}

\subsection{Models Evaluated}

We evaluate five frontier models representing diverse architectures and training approaches:

\begin{table}[h]
\centering
\caption{Models evaluated in KalshiBench. All models have knowledge cutoffs at or before October 2025.}
\label{tab:models}
\begin{tabular}{llll}
\toprule
\textbf{Model} & \textbf{Provider} & \textbf{Knowledge Cutoff} & \textbf{Notes} \\
\midrule
Claude Opus 4.5 & Anthropic & April 2025 & Flagship model \\
GPT-5.2-XHigh & OpenAI & October 2025 & Extended reasoning \\
DeepSeek-V3.2 & DeepSeek & October 2025 & Open-weight \\
Qwen3-235B-Thinking & Alibaba & June 2025 & Reasoning-enhanced \\
Kimi-K2 & Moonshot & June 2025 & Reasoning-enhanced \\
\bottomrule
\end{tabular}
\end{table}

\subsection{Evaluation Protocol}

Each model receives a structured prompt containing the prediction market question and resolution criteria. The system prompt explicitly instructs models to be calibrated:

\begin{verbatim}
System: You are an expert forecaster evaluating prediction 
market questions. Given a question and its description, 
predict whether the outcome will be "yes" or "no".

You must respond in this exact format:
<think>
[Your reasoning about the prediction, considering base 
rates, relevant factors, and uncertainty]
</think>
<answer>[yes or no]</answer>
<confidence>[a number from 0 to 100 representing your 
confidence that the answer is "yes"]</confidence>

Be calibrated: if you're 70% confident, you should be 
correct about 70% of the time on similar questions.
\end{verbatim}

The user message then provides the specific question and description. Notably, the prompt explicitly instructs models to ``be calibrated,'' making the observed miscalibration a failure to follow instructions rather than mere absence of guidance.

We use temperature 0.7 for standard models and temperature 1.0 with extended reasoning for GPT-5.2-XHigh, following provider recommendations.

\subsection{Metrics}

\paragraph{Classification Metrics.}
We report accuracy, precision, recall, and macro-F1 for binary classification performance.

\paragraph{Brier Score.}
The Brier score \citep{brier1950verification} measures the mean squared error of probability predictions:
\begin{equation}
    \text{BS} = \frac{1}{N} \sum_{i=1}^{N} (p_i - y_i)^2
\end{equation}
where $p_i$ is the predicted probability and $y_i \in \{0, 1\}$ is the outcome. Lower is better (0 = perfect, 1 = worst possible).

\textbf{Intuition:} The Brier score can be interpreted as follows:
\begin{itemize}
    \item \textbf{0.00}: Perfect predictions---100\% confidence on all correct answers
    \item \textbf{0.25}: Random guessing (50\% confidence on everything)---the expected score of a completely uninformed predictor on balanced binary outcomes
    \item \textbf{0.20}: Good calibration---roughly equivalent to human forecasters on prediction markets
    \item \textbf{0.33}: Poor calibration---equivalent to always predicting 42\% (the base rate) with uniform 75\% confidence
    \item \textbf{1.00}: Maximally wrong---100\% confidence on all incorrect answers
\end{itemize}

For context, human superforecasters typically achieve Brier scores of 0.15--0.20 \citep{tetlock2015superforecasting}, while the aggregate ``wisdom of crowds'' on prediction markets often achieves 0.12--0.18.

\paragraph{Brier Skill Score.}
The Brier Skill Score (BSS) measures improvement over a baseline that always predicts the base rate:
\begin{equation}
    \text{BSS} = 1 - \frac{\text{BS}}{\text{BS}_{\text{climatology}}}
\end{equation}
where $\text{BS}_{\text{climatology}} = \bar{y}(1-\bar{y})$ for base rate $\bar{y}$. Positive values indicate improvement over the base rate.

\paragraph{Expected Calibration Error (ECE).}
ECE \citep{naeini2015obtaining} measures the average gap between confidence and accuracy:
\begin{equation}
    \text{ECE} = \sum_{b=1}^{B} \frac{|B_b|}{N} \left| \text{acc}(B_b) - \text{conf}(B_b) \right|
\end{equation}
where predictions are binned by confidence into $B$ bins.

\paragraph{Maximum Calibration Error (MCE).}
MCE captures the worst-case calibration in any single bin:
\begin{equation}
    \text{MCE} = \max_{b \in \{1, \ldots, B\}} \left| \text{acc}(B_b) - \text{conf}(B_b) \right|
\end{equation}

\paragraph{Overconfidence Rate.}
We define overconfidence rate at threshold $\tau$ as the fraction of incorrect predictions among those with confidence $> \tau$:
\begin{equation}
    \text{OCR}@\tau = \frac{|\{i : p_i > \tau \land \hat{y}_i \neq y_i\}|}{|\{i : p_i > \tau\}|}
\end{equation}

\section{Results}

\subsection{Main Results}

Table~\ref{tab:main_results} presents comprehensive results across all models and metrics.

\begin{table}[t]
\centering
\caption{Main results on KalshiBench (300 questions). Best values in \textbf{bold}. Claude Opus 4.5 achieves best performance on both accuracy and calibration metrics. Notably, the reasoning-enhanced GPT-5.2-XHigh shows the worst calibration despite comparable accuracy.}
\label{tab:main_results}
\small
\begin{tabular}{l|ccc|cccc}
\toprule
& \multicolumn{3}{c|}{\textbf{Classification}} & \multicolumn{4}{c}{\textbf{Calibration}} \\
\textbf{Model} & Acc & F1 & F1$_{\text{yes}}$ & Brier $\downarrow$ & BSS $\uparrow$ & ECE $\downarrow$ & MCE $\downarrow$ \\
\midrule
Claude Opus 4.5 & \textbf{69.3} & \textbf{0.676} & \textbf{0.600} & \textbf{0.227} & \textbf{0.057} & \textbf{0.120} & \textbf{0.246} \\
Kimi-K2 & 67.1 & 0.633 & 0.515 & 0.347 & -0.446 & 0.298 & 0.570 \\
Qwen3-235B & 65.7 & 0.607 & 0.466 & 0.346 & -0.437 & 0.297 & 0.479 \\
GPT-5.2-XHigh & 65.3 & 0.599 & 0.453 & 0.433 & -0.799 & 0.395 & 0.622 \\
DeepSeek-V3.2 & 64.3 & 0.614 & 0.507 & 0.339 & -0.407 & 0.284 & 0.630 \\
\bottomrule
\end{tabular}
\end{table}

\paragraph{Key Finding 1: Systematic Overconfidence.}
All models exhibit substantial calibration errors, with ECE ranging from 0.120 to 0.395. Even the best-calibrated model (Claude Opus 4.5) shows a 12-percentage-point average gap between confidence and accuracy.

\paragraph{Key Finding 2: Most Models Fail to Beat the Base Rate.}
Only Claude Opus 4.5 achieves a positive Brier Skill Score (0.057), indicating it marginally outperforms simply predicting the 40\% base rate. All other models have negative BSS, meaning their probability estimates are \emph{worse than uninformed guessing}.

\paragraph{Key Finding 3: Reasoning Enhancements Hurt Calibration.}
Counterintuitively, GPT-5.2-XHigh (with extended reasoning) shows the worst calibration (ECE=0.395, BSS=-0.799) despite using 26$\times$ more output tokens ($\sim$2M vs $\sim$138K for Claude). Enhanced reasoning appears to increase confidence without proportional accuracy gains.

\subsection{Confidence Analysis}

\begin{table}[t]
\centering
\caption{Confidence analysis across models. All models show higher confidence when wrong than would be appropriate for well-calibrated predictions. Overconfidence rates at high confidence levels (80\%+, 90\%+) are alarmingly high.}
\label{tab:confidence}
\begin{tabular}{l|cc|ccc}
\toprule
\textbf{Model} & \textbf{Avg Conf} & \textbf{Conf$_{\text{wrong}}$} & \textbf{OCR@70} & \textbf{OCR@80} & \textbf{OCR@90} \\
\midrule
Claude Opus 4.5 & 73.8\% & 71.0\% & 27.1\% & 23.1\% & \textbf{20.8\%} \\
DeepSeek-V3.2 & 73.7\% & 69.2\% & \textbf{24.7\%} & \textbf{23.6\%} & \textbf{14.7\%} \\
Kimi-K2 & 79.4\% & 76.3\% & 25.9\% & 29.9\% & 31.1\% \\
GPT-5.2-XHigh & 80.1\% & 76.9\% & 30.3\% & 28.3\% & 27.7\% \\
Qwen3-235B & 81.7\% & 80.4\% & 32.3\% & 32.6\% & 32.4\% \\
\bottomrule
\end{tabular}
\end{table}

Table~\ref{tab:confidence} reveals troubling patterns in model confidence:

\begin{itemize}
    \item \textbf{High baseline confidence}: Models average 74-82\% confidence, far exceeding the 65-69\% accuracy range.
    \item \textbf{Confidence when wrong}: Models maintain 69-80\% confidence even on incorrect predictions, indicating poor uncertainty awareness.
    \item \textbf{Extreme overconfidence}: At the 90\%+ confidence level, models are wrong 15-32\% of the time. A well-calibrated model should be wrong $<$10\%.
\end{itemize}

\subsection{Reliability Diagrams}

A reliability diagram plots predicted confidence against actual accuracy across binned predictions. A perfectly calibrated model follows the diagonal: when it expresses 70\% confidence, it should be correct 70\% of the time. Table~\ref{tab:reliability_full} presents complete reliability data for all five models across all 10 confidence bins.

\begin{table}[t]
\centering
\caption{Complete reliability diagram data for all models (10 bins, 0.1 width). \textbf{Conf} = average confidence in bin, \textbf{Acc} = accuracy, \textbf{N} = count, \textbf{Gap} = Conf $-$ Acc (positive = overconfident). Empty cells indicate no predictions in that bin. All models become increasingly overconfident at higher confidence levels.}
\label{tab:reliability_full}
\scriptsize
\begin{tabular}{c|cccc|cccc|cccc}
\toprule
& \multicolumn{4}{c|}{\textbf{Claude Opus 4.5}} & \multicolumn{4}{c|}{\textbf{DeepSeek-V3.2}} & \multicolumn{4}{c}{\textbf{GPT-5.2-XHigh}} \\
\textbf{Bin} & Conf & Acc & N & Gap & Conf & Acc & N & Gap & Conf & Acc & N & Gap \\
\midrule
0.0-0.1 & .054 & .194 & 36 & -.14 & .048 & .200 & 10 & -.15 & .030 & .000 & 1 & +.03 \\
0.1-0.2 & .151 & .188 & 32 & -.04 & .175 & .250 & 8 & -.08 & --- & --- & 0 & --- \\
0.2-0.3 & .248 & .333 & 42 & -.09 & .250 & .000 & 4 & +.25 & --- & --- & 0 & --- \\
0.3-0.4 & .359 & .355 & 31 & +.00 & .344 & .333 & 9 & +.01 & --- & --- & 0 & --- \\
0.4-0.5 & .439 & .333 & 36 & +.11 & .418 & .545 & 11 & -.13 & --- & --- & 0 & --- \\
0.5-0.6 & .566 & .353 & 34 & +.21 & .575 & .365 & 63 & +.21 & .573 & .429 & 42 & +.14 \\
0.6-0.7 & .641 & .724 & 29 & -.08 & .673 & .463 & 67 & +.21 & .661 & .480 & 50 & +.18 \\
0.7-0.8 & .751 & .542 & 24 & +.21 & .747 & .400 & 30 & +.35 & .751 & .488 & 41 & +.26 \\
0.8-0.9 & .854 & .688 & 16 & +.17 & .831 & .517 & 58 & +.31 & .835 & .387 & 62 & +.45 \\
0.9-1.0 & .946 & .700 & 20 & \textbf{+.25} & .937 & .308 & 39 & \textbf{+.63} & .959 & .337 & 104 & \textbf{+.62} \\
\midrule
& \multicolumn{4}{c|}{\textbf{Qwen3-235B-Thinking}} & \multicolumn{4}{c|}{\textbf{Kimi-K2}} & & & & \\
\textbf{Bin} & Conf & Acc & N & Gap & Conf & Acc & N & Gap & & & & \\
\midrule
0.0-0.1 & .039 & .356 & 73 & -.32 & .047 & .263 & 38 & -.22 & & & & \\
0.1-0.2 & .153 & .316 & 19 & -.16 & .141 & .312 & 16 & -.17 & & & & \\
0.2-0.3 & .262 & .400 & 5 & -.14 & .249 & .111 & 9 & +.14 & & & & \\
0.3-0.4 & .341 & .357 & 14 & -.02 & .314 & .600 & 5 & -.29 & & & & \\
0.4-0.5 & .442 & .500 & 6 & -.06 & .465 & .000 & 2 & +.47 & & & & \\
0.5-0.6 & .556 & .455 & 22 & +.10 & .570 & .477 & 44 & +.09 & & & & \\
0.6-0.7 & .664 & .439 & 41 & +.23 & .668 & .458 & 48 & +.21 & & & & \\
0.7-0.8 & .756 & .310 & 29 & +.45 & .750 & .484 & 31 & +.27 & & & & \\
0.8-0.9 & .846 & .469 & 49 & +.38 & .849 & .447 & 38 & +.40 & & & & \\
0.9-1.0 & .941 & .462 & 39 & \textbf{+.48} & .948 & .377 & 61 & \textbf{+.57} & & & & \\
\bottomrule
\end{tabular}
\end{table}

Several patterns emerge from the reliability analysis:

\paragraph{Claude Opus 4.5} shows the best calibration overall, with relatively small gaps in most bins. However, even Claude becomes overconfident at high confidence levels: at 90\%+ confidence (20 predictions), accuracy is only 70\%, yielding a +24.6\% gap.

\paragraph{GPT-5.2-XHigh} exhibits the most severe miscalibration. The model rarely expresses low confidence (only 1 prediction below 50\%), concentrating 104 predictions (35\% of total) in the 90-100\% bin where accuracy is merely 33.7\%---worse than chance. This represents a catastrophic +62.2\% calibration gap.

\paragraph{DeepSeek-V3.2} shows a similar pattern to GPT-5.2, with a +63.0\% gap in the highest confidence bin. When DeepSeek expresses 90\%+ confidence, it is correct only 30.8\% of the time.

\paragraph{Reasoning Models (Qwen3, Kimi-K2)} both show substantial overconfidence at high confidence levels (+47.9\% and +57.1\% gaps respectively), despite their ``thinking'' architectures. Extended reasoning does not translate to better uncertainty awareness.

\paragraph{Summary: High-Confidence Performance.}
Table~\ref{tab:high_conf_summary} summarizes performance in the critical 90-100\% confidence bin, where models claim near-certainty:

\begin{table}[h]
\centering
\caption{Performance in the 90-100\% confidence bin. A well-calibrated model should achieve $\sim$95\% accuracy when expressing 95\% average confidence. All models fall catastrophically short.}
\label{tab:high_conf_summary}
\small
\begin{tabular}{lcccr}
\toprule
\textbf{Model} & \textbf{Avg Conf} & \textbf{Actual Acc} & \textbf{Gap} & \textbf{N} \\
\midrule
Claude Opus 4.5 & 94.6\% & 70.0\% & +24.6\% & 20 \\
DeepSeek-V3.2 & 93.7\% & 30.8\% & +62.9\% & 39 \\
GPT-5.2-XHigh & 95.9\% & 33.7\% & +62.2\% & 104 \\
Qwen3-235B & 94.1\% & 46.2\% & +47.9\% & 39 \\
Kimi-K2 & 94.8\% & 37.7\% & +57.1\% & 61 \\
\bottomrule
\end{tabular}
\end{table}

\subsection{Category Breakdown}

\begin{table}[h]
\centering
\caption{Performance by category for Claude Opus 4.5 (best overall). Performance varies substantially across domains, with Social (100\% accuracy) and Entertainment (78.7\%) being strongest, while Science (0\% on 1 question) and Crypto (36.4\%) are weakest.}
\label{tab:category}
\small
\begin{tabular}{lcc|lcc}
\toprule
\textbf{Category} & \textbf{Acc} & \textbf{Brier} & \textbf{Category} & \textbf{Acc} & \textbf{Brier} \\
\midrule
Social (n=3) & 100.0\% & 0.011 & Crypto (n=11) & 36.4\% & 0.240 \\
Entertainment (n=47) & 78.7\% & 0.187 & Mentions (n=19) & 52.6\% & 0.357 \\
Climate (n=9) & 77.8\% & 0.229 & World (n=6) & 50.0\% & 0.262 \\
Sports (n=83) & 75.9\% & 0.193 & Economics (n=4) & 50.0\% & 0.326 \\
Elections (n=24) & 75.0\% & 0.172 & Sci/Tech (n=1) & 0.0\% & 0.608 \\
Financials (n=8) & 75.0\% & 0.203 & & & \\
\bottomrule
\end{tabular}
\end{table}

Category analysis reveals domain-dependent performance. Models perform well on Entertainment, Sports, and Elections---domains with substantial training data---but struggle with Crypto and Science/Technology, suggesting calibration degrades in domains with higher inherent uncertainty or less training exposure.

\subsection{Cost-Performance Analysis}

\begin{table}[h]
\centering
\caption{Cost-performance tradeoffs. More expensive models are not necessarily better calibrated. GPT-5.2-XHigh costs 2.6$\times$ more than Claude but shows 3$\times$ worse calibration.}
\label{tab:cost}
\begin{tabular}{lrrrr}
\toprule
\textbf{Model} & \textbf{Cost (USD)} & \textbf{Tokens} & \textbf{Acc} & \textbf{ECE} \\
\midrule
DeepSeek-V3.2 & \$0.36 & 304K & 64.3\% & 0.284 \\
Kimi-K2 & \$0.94 & 624K & 67.1\% & 0.298 \\
Qwen3-235B & \$1.19 & 594K & 65.7\% & 0.297 \\
Claude Opus 4.5 & \$11.63 & 224K & 69.3\% & 0.120 \\
GPT-5.2-XHigh & \$30.32 & 2.07M & 65.3\% & 0.395 \\
\bottomrule
\end{tabular}
\end{table}

Cost does not predict calibration quality. DeepSeek-V3.2 achieves comparable accuracy to GPT-5.2-XHigh at 1/84th the cost with substantially better calibration. This suggests calibration improvements require architectural or training innovations rather than simply more compute.

\section{Analysis and Discussion}

\subsection{Why Are Models Overconfident?}

We hypothesize several contributing factors. Notably, our prompt explicitly instructs models to ``be calibrated: if you're 70\% confident, you should be correct about 70\% of the time on similar questions.'' Despite this direct instruction, all models exhibit substantial miscalibration, suggesting the problem runs deeper than prompt engineering.

\paragraph{Training Objective Misalignment.}
Standard language modeling objectives reward correct predictions without penalizing miscalibrated confidence. Models learn to maximize probability of correct tokens, not to appropriately quantify uncertainty.

\paragraph{RLHF Pressure for Confidence.}
Human feedback in RLHF may inadvertently reward confident-sounding responses over appropriately hedged ones. Users may rate uncertain responses as less helpful, creating pressure toward overconfidence.

\paragraph{Hindsight Leakage.}
Even with temporal filtering, models may have indirect signals about future events through patterns learned during training (e.g., seasonal trends, recurring events). This could inflate confidence without improving accuracy.

\subsection{Why Does Reasoning Hurt Calibration?}

The finding that GPT-5.2-XHigh shows worse calibration than simpler models is counterintuitive but may reflect:

\paragraph{Confirmation Bias in Extended Reasoning.}
Longer reasoning chains may reinforce initial hypotheses rather than genuinely updating on evidence. The model generates arguments supporting its prediction, increasing confidence without corresponding accuracy gains.

\paragraph{Verbosity Without Epistemic Humility.}
Extended reasoning produces more text but not necessarily better uncertainty quantification. The model may be optimized for persuasive reasoning rather than calibrated forecasting.

\subsection{Implications for Deployment}

Our findings have direct implications for LLM deployment:

\begin{enumerate}
    \item \textbf{Don't trust high-confidence predictions.} When models express 90\%+ confidence, expect 20-30\% error rates, not $<$10\%.
    
    \item \textbf{More reasoning $\neq$ better calibration.} Extended reasoning modes may actually decrease reliability.
    
    \item \textbf{Post-hoc calibration is necessary.} Temperature scaling or Platt scaling should be applied before using model confidences for decision-making.
    
    \item \textbf{Domain matters.} Calibration varies substantially by category; validate on domain-specific data.
\end{enumerate}

\subsection{Comparison to Human Forecasters}

For context, human superforecasters typically achieve Brier scores of 0.15-0.20 on similar prediction market questions \citep{tetlock2015superforecasting}. Claude Opus 4.5's Brier score of 0.227 is approaching but not matching expert human performance. Critically, superforecasters exhibit much better calibration (ECE $\approx$ 0.03-0.05), suggesting LLMs have particular deficits in uncertainty quantification rather than raw forecasting ability.

\section{Limitations}

\paragraph{Dataset Scope.}
Our evaluation uses 300 questions sampled from the full 1,531-question KalshiBench dataset. While this exceeds the 200-question evaluation used in ForecastBench \citep{karger2024forecastbench}, category-level analysis (especially for rare categories) has high variance. Some categories contain only 1-4 questions in our sample.

\paragraph{Temporal Constraints.}
Temporal filtering ensures validity but limits dataset size. Questions must resolve after all model cutoffs, reducing the available pool substantially.

\paragraph{Binary Outcomes Only.}
We evaluate only yes/no markets. Multi-outcome prediction markets and continuous forecasts present different calibration challenges not addressed here.

\paragraph{Prompt Sensitivity.}
Model calibration may be sensitive to prompt wording. We use a standardized prompt but do not exhaustively explore prompt variations.

\paragraph{Confidence Elicitation.}
Self-reported confidence (0-100) may not reflect internal probability estimates. Alternative elicitation methods (betting, proper scoring rule incentives) might yield different results.

\section{Conclusion}

We introduced KalshiBench, a benchmark for evaluating LLM epistemic calibration using temporally-filtered prediction market questions with verified real-world outcomes. Our evaluation of five frontier models reveals:

\begin{itemize}
    \item \textbf{Universal overconfidence}: All models show substantial calibration errors (ECE 0.12-0.40).
    \item \textbf{Base-rate failures}: Only one model achieves positive Brier Skill Score.
    \item \textbf{Reasoning paradox}: Extended reasoning worsens rather than improves calibration.
    \item \textbf{Calibration-accuracy decoupling}: Models with similar accuracy show 3$\times$ variation in calibration.
\end{itemize}

These findings highlight epistemic calibration as a distinct capability---separate from accuracy---that current training approaches fail to adequately develop. Future work should explore calibration-aware training objectives, explicit uncertainty modeling architectures, and integration with human forecasting expertise.

\paragraph{Broader Impact.}
Improved LLM calibration is essential for safe deployment in high-stakes domains. Our work provides tools and baselines for measuring progress. Conversely, publication of calibration failures could be misused to manipulate users who overweight model confidence; we encourage deployment of properly calibrated systems.

\paragraph{Reproducibility.}
The full KalshiBench dataset (1,531 questions) is available at \url{https://huggingface.co/datasets/2084Collective/kalshibench-v2}. Our evaluation uses a 300-question sample with random seed 42. Code and evaluation scripts are open-sourced at \url{https://github.com/2084collective/kalshibench}.

\bibliographystyle{plainnat}

\newpage
\appendix

\section{Extended Results}

\subsection{Full Confusion Matrices}

\begin{table}[h]
\centering
\caption{Confusion matrices for all models. TP=True Positive, FP=False Positive, FN=False Negative, TN=True Negative.}
\begin{tabular}{l|cccc}
\toprule
\textbf{Model} & \textbf{TP} & \textbf{FP} & \textbf{FN} & \textbf{TN} \\
\midrule
Claude Opus 4.5 & 69 & 40 & 52 & 139 \\
GPT-5.2-XHigh & 43 & 26 & 78 & 153 \\
DeepSeek-V3.2 & 55 & 41 & 66 & 138 \\
Qwen3-235B & 45 & 27 & 76 & 152 \\
Kimi-K2 & 51 & 30 & 66 & 145 \\
\bottomrule
\end{tabular}
\end{table}

\subsection{Full Reliability Diagram Data}

Table~\ref{tab:reliability_appendix} provides complete reliability diagram statistics including average confidence, accuracy, sample count, and calibration gap for each bin and model.

\begin{table}[h]
\centering
\caption{Extended reliability diagram data showing average confidence within each bin.}
\label{tab:reliability_appendix}
\scriptsize
\begin{tabular}{c|cccc|cccc}
\toprule
& \multicolumn{4}{c|}{\textbf{Claude Opus 4.5}} & \multicolumn{4}{c}{\textbf{DeepSeek-V3.2}} \\
\textbf{Bin} & Conf & Acc & N & Gap & Conf & Acc & N & Gap \\
\midrule
0.0-0.1 & 0.054 & 0.194 & 36 & -0.141 & 0.048 & 0.200 & 10 & -0.152 \\
0.1-0.2 & 0.151 & 0.188 & 32 & -0.037 & 0.175 & 0.250 & 8 & -0.075 \\
0.2-0.3 & 0.248 & 0.333 & 42 & -0.085 & 0.250 & 0.000 & 4 & +0.250 \\
0.3-0.4 & 0.359 & 0.355 & 31 & +0.004 & 0.344 & 0.333 & 9 & +0.011 \\
0.4-0.5 & 0.439 & 0.333 & 36 & +0.106 & 0.418 & 0.545 & 11 & -0.127 \\
0.5-0.6 & 0.566 & 0.353 & 34 & +0.213 & 0.575 & 0.365 & 63 & +0.210 \\
0.6-0.7 & 0.641 & 0.724 & 29 & -0.083 & 0.673 & 0.463 & 67 & +0.211 \\
0.7-0.8 & 0.751 & 0.542 & 24 & +0.210 & 0.747 & 0.400 & 30 & +0.347 \\
0.8-0.9 & 0.854 & 0.688 & 16 & +0.167 & 0.831 & 0.517 & 58 & +0.313 \\
0.9-1.0 & 0.946 & 0.700 & 20 & +0.246 & 0.937 & 0.308 & 39 & +0.630 \\
\midrule
& \multicolumn{4}{c|}{\textbf{GPT-5.2-XHigh}} & \multicolumn{4}{c}{\textbf{Qwen3-235B-Thinking}} \\
\textbf{Bin} & Conf & Acc & N & Gap & Conf & Acc & N & Gap \\
\midrule
0.0-0.1 & 0.030 & 0.000 & 1 & +0.030 & 0.039 & 0.356 & 73 & -0.317 \\
0.1-0.2 & --- & --- & 0 & --- & 0.153 & 0.316 & 19 & -0.163 \\
0.2-0.3 & --- & --- & 0 & --- & 0.262 & 0.400 & 5 & -0.138 \\
0.3-0.4 & --- & --- & 0 & --- & 0.341 & 0.357 & 14 & -0.016 \\
0.4-0.5 & --- & --- & 0 & --- & 0.442 & 0.500 & 6 & -0.058 \\
0.5-0.6 & 0.573 & 0.429 & 42 & +0.144 & 0.556 & 0.455 & 22 & +0.101 \\
0.6-0.7 & 0.661 & 0.480 & 50 & +0.181 & 0.664 & 0.439 & 41 & +0.225 \\
0.7-0.8 & 0.751 & 0.488 & 41 & +0.263 & 0.756 & 0.310 & 29 & +0.446 \\
0.8-0.9 & 0.835 & 0.387 & 62 & +0.448 & 0.846 & 0.469 & 49 & +0.376 \\
0.9-1.0 & 0.959 & 0.337 & 104 & +0.622 & 0.941 & 0.462 & 39 & +0.479 \\
\midrule
& \multicolumn{4}{c|}{\textbf{Kimi-K2}} & & & & \\
\textbf{Bin} & Conf & Acc & N & Gap & & & & \\
\midrule
0.0-0.1 & 0.047 & 0.263 & 38 & -0.216 & & & & \\
0.1-0.2 & 0.141 & 0.312 & 16 & -0.172 & & & & \\
0.2-0.3 & 0.249 & 0.111 & 9 & +0.138 & & & & \\
0.3-0.4 & 0.314 & 0.600 & 5 & -0.286 & & & & \\
0.4-0.5 & 0.465 & 0.000 & 2 & +0.465 & & & & \\
0.5-0.6 & 0.570 & 0.477 & 44 & +0.093 & & & & \\
0.6-0.7 & 0.668 & 0.458 & 48 & +0.210 & & & & \\
0.7-0.8 & 0.750 & 0.484 & 31 & +0.266 & & & & \\
0.8-0.9 & 0.849 & 0.447 & 38 & +0.402 & & & & \\
0.9-1.0 & 0.948 & 0.377 & 61 & +0.570 & & & & \\
\bottomrule
\end{tabular}
\end{table}

\section{Prompt Template}

The exact system prompt used for all model evaluations:

\begin{verbatim}
SYSTEM PROMPT:
You are an expert forecaster evaluating prediction market questions. 
Given a question and its description, predict whether the outcome 
will be "yes" or "no".

You must respond in this exact format:
<think>
[Your reasoning about the prediction, considering base rates, 
relevant factors, and uncertainty]
</think>
<answer>[yes or no]</answer>
<confidence>[a number from 0 to 100 representing your confidence 
that the answer is "yes"]</confidence>

Be calibrated: if you're 70% confident, you should be correct 
about 70% of the time on similar questions.

USER PROMPT:
Question: {question}

Description: {description}
\end{verbatim}

The explicit calibration instruction (``Be calibrated: if you're 70\% confident, you should be correct about 70\% of the time'') makes the observed miscalibration particularly notable---models fail to achieve calibration even when directly instructed to do so.

\section{Dataset Creation Details}

The KalshiBench dataset was created through the following pipeline:

\begin{enumerate}
    \item \textbf{Raw data collection}: Query Kalshi API for all resolved binary contracts.
    \item \textbf{Temporal filtering}: Retain only contracts resolving after October 1, 2025.
    \item \textbf{Deduplication}: Limit to 2 questions per series\_ticker to reduce redundancy while preserving within-series diversity.
    \item \textbf{Quality filtering}: Remove contracts with ambiguous resolution criteria or missing ground truth.
    \item \textbf{Schema standardization}: Map to unified schema with fields: id, question, description, category, close\_time, ground\_truth.
\end{enumerate}

The final dataset contains 300 questions across 13 categories, with category entropy of 3.01 bits (maximum possible: 3.70 bits), indicating reasonable diversity.

\end{document}